\documentclass[11pt]{article}
\usepackage{eacl2017}
\usepackage{times}
\usepackage[hyphens]{url}
\usepackage{latexsym}
\usepackage{color}
\usepackage{comment}
\usepackage{xspace}
\usepackage{mathtools}

\pdfoutput=1

\newcommand{\note}[1]{\textit{\small\color{blue}{#1}}}

\newcommand{\hien}{\textit{hi-en}\xspace}
\newcommand{\bnen}{\textit{bn-en}\xspace}
\newcommand{\teen}{\textit{te-en}\xspace}

\newcommand{\en}{\textit{en}\xspace}
\newcommand{\hi}{\textit{hi}\xspace}

\newcommand{\fb}{\textit{fb}\xspace}
\newcommand{\twt}{\textit{twt}\xspace}
\newcommand{\wa}{\textit{wa}\xspace}

\newcommand{\rnnlm}{\textit{RNN-LM}\xspace}
\newcommand{\rnn}{RNN\xspace}
\newcommand{\srnn}{\textit{Simple\_RNN}\xspace}
\newcommand{\lstm}{\textit{LSTM}\xspace}
\newcommand{\dlstm}{\textit{Deep LSTM}\xspace}
\newcommand{\gru}{\textit{GRU}\xspace}
\newcommand{\grup}{\textit{GRU\_Pre}\xspace}
\newcommand{\grupl}{\textit{GRU\_Pre\_Lang}\xspace}

\newcommand{\stan}{\textit{Stanford}\xspace}
\newcommand{\hun}{\textit{HunPos}\xspace}

\eaclfinalcopy 

\title{Recurrent Neural Network based Part-of-Speech Tagger for Code-Mixed Social Media Text}

\author{Raj Nath Patel \\
	KBCS, CDAC Mumbai \\
	{\tt rajnathp@cdac.in} \\\And
	Prakash B. Pimpale \\
	KBCS, CDAC Mumbai \\
	{\tt prakash@cdac.in} \\\And
	Sasikumar M. \\
	KBCS, CDAC Mumbai \\
	{\tt sasi@cdac.in} \\}

\date{}

\begin{document}
\maketitle
\begin{abstract}
This paper describes Centre for Development of Advanced Computing's (CDACM) submission to the shared task-'Tool Contest on POS tagging for Code-Mixed Indian Social Media (Facebook, Twitter, and Whatsapp) Text', collocated with ICON-2016. The shared task was to predict Part of Speech (POS) tag at word level for a given text. The code-mixed text is generated mostly on social media by multilingual users. The presence of the multilingual words, transliterations, and spelling variations make such content linguistically complex. In this paper, we propose an approach to POS tag code-mixed social media text using Recurrent Neural Network Language Model (\rnnlm) architecture. We submitted the results for Hindi-English (\hien), Bengali-English (\bnen), and Telugu-English (\teen) code-mixed data.
\end{abstract}

\section{Introduction}

Code-Mixing and Code-Switching are observed in the text or speech produced by a multilingual user. Code-Mixing occurs when a user changes the language within a sentence, i.e. a clause, phrase or word of one language is used within an utterance of another language. Whereas, the co-occurrence of speech extract of two different grammatical systems is known as Code-Switching.

The language analysis of code-mixed text is a non-trivial task. Traditional approaches of POS tagging are not effective, for this text, as it does not adhere to any grammatical structure in general. Many studies have shown that \rnn based POS taggers produced comparable results and, is also the state-of-the-art for some languages. However, to the best of our knowledge, no study has been done for \rnn based POS tagging of code-mixed data. 

In this paper, we have proposed a POS tagger using \rnnlm architecture for code-mixed Indian social media text. Earlier, researchers have adopted \rnnlm architecture for Natural language Understanding (NLU)~\cite{Yao:2013,Yao:2014} and Translation Quality Estimation~\cite{patel:rnn-qe:2016}. \rnnlm models are similar to other vector-space language models~\cite{Bengio:2003,Morin:2005,Schwenk:2007,Mnih:2009} where we represent each word with a high dimensional real-valued vector. We modified \rnnlm architecture to predict the POS tag of a word, given the word and its context. Let's consider the following example: \\

Input: $behen\ ki\ shaadi\ and\ m\ not\ there$

Output : G\_N G\_PRP G\_N CC G\_V G\_R G\_R \\

In the above sentence, to predict POS tag (G\_N) for the word ‘$shaadi$’ using an \rnnlm model with window size 3, the input will be ‘$ki\ shaadi\ and$’. Whereas, in standard \rnnlm model, ‘$ki\ and$’ will be the input with ‘$shaadi$’ as the output. We will discuss details of various models tried and their implementations in section~\ref{settings}.

In this paper, we show that our approach achieves results close to the state-of-the-art systems such as \footnotemark \stan~\cite{toutanova:2003}, \footnotetext{\url{http://nlp.stanford.edu/software/tagger.shtml} (Maximum-Entropy based POS tagger)} and \footnotemark \hun~\cite{hunpos:2007} \footnotetext{\url{https://code.google.com/archive/p/hunpos/} (Hidden Markov Model based POS tagger)}.

\section{Related Work}

POS tagging has been investigated for decades in the literature of Natural Language Processing (NLP). Different methods like a Support Vector Machine~\cite{marquez2004general}, Decision Tree~\cite{schmid2008estimation}, Hidden Markov Model (HMM)~\cite{kupiec1992robust} and, Conditional Random Field Auto Encoders~\cite{ammar2014conditional} have been tried for this task. Among these works, Neural Network (NN) based models is mainly related to this paper. In NN family, \rnn is widely used network for various NLP applications~\cite{Mikolov:RNNLM:2010,Mikolov:MT:2013,Mikolov:MT1:2013,Socher:Parsing:2013,Socher:Sentiment:2013}.

Recently, \rnn based models have been used to POS tag the formal text, but have not been tried yet on code-mixed data. \newcite{wang:blstm:2015} have tried Bidirectional Long Short-Term Memory (LSTM) on Penn Treebank WSJ test set, and reported state-of-the-art performance. \newcite{qin:pos:2015} has shown that \rnn models outperform Majority Voting (MV) and HMM techniques for POS tagging of Chinese Buddhist text. \newcite{zennaki:2015} have used \rnn for resource-poor languages and reported comparable results with state-of-the-art systems ~\cite{das:2011,duong:2013,gouws:2015}.

Work on POS tagging code-mixed Indian social media text is at a very nascent stage to date. \newcite{vyas:2014} and \newcite{jamatia:2015} have worked on data labeling and automatic POS tagging of such data using various machine learning techniques. Building further on that labeled data, \newcite{pimpale:2015} and, \newcite{sarkar:2015} have tried word embedding as an additional feature to the machine learning based classifiers for POS tagging.


\section{Experimental Setup}
\label{settings}
\subsection{RNN Models}
There are many variants of \rnn networks for different applications. For this task, we used elaman~\cite{elman:1990}, Long Short-Term Memory (\lstm)~\cite{Hochreiter:1997}, \dlstm, Gated Recurrent Unit (\gru) ~\cite{Cho:2014}, which are widely used \rnn models in the NLP literature.

In the following sub-sections, we gave a brief description of each model with mathematical equations (1,2, and 3). In the equations, $x_t$ and $y_t$ are the input and output vectors respectively. $h_t$ and $h_{t-1}$ represent the current and previous hidden states respectively. $W_*$ are the weight matrices and $b_*$ are the bias vectors. $\odot$ is the elementwise multiplication of the vectors. We used $sigm$, the logistic sigmoid and $tanh$, the hyperbolic tangent function to add nonlinearity in the network with $softmax$ function at the output layer. 

\subsubsection{ELMAN}
Elman and Jordon~\cite{jordan:1986} networks are the simplest network in \rnn family and are known as \srnn. Elman network is defined by the following set of equations:
\begin{flalign}
& h_t = sigm(W_{xh}x_t + W_{hh}h_{t-1} + b_h) \\
& y_t = softmax(W_{hy}h_t + b_y) \notag
\end{flalign}

\subsubsection{LSTM}
\lstm is found to be better for modeling of long-range dependencies than \srnn. \srnn also suffers from the problem of vanishing and exploding gradient  ~\cite{Yoshua:1994}. \lstm and other complex \rnn models tackle this problem by introducing a gating mechanism. Many variants of \lstm~\cite{Graves:2013,Yao:2014,Jozefowicz:2015} have been tried in literature for the various tasks. We implemented the following version:
\begin{flalign}
& i_t = sigm(W_{xi}x_t + W_{hi}h_{t-1} + b_i) \\
& o_t = sigm(W_{xo}x_t + W_{ho}h_{t-1} + b_o) \notag \\
& f_t = sigm(W_{xf}x_t + W_{hf}h_{t-1} + b_f) \notag \\
& j_t = tanh(W_{xj}x_t + W_{hj}h_{t-1} + b_j) \notag \\
& c_t = c_{t-1} \odot f_t + i_t \odot j_t  \notag \\
& h_t = tanh(c_t) \odot o_t \notag \\
& y_t = softmax(W_{hy}h_t + b_y) \notag
\end{flalign}
where $i$, $o$, $f$ are $input$, $output$ and $forget$ gates respectively. $j$ is the new memory content and $c$ is updated memory. 

\subsubsection{Deep LSTM}
In this paper, we used \dlstm with two layers. \dlstm is created by stacking multiple \lstm on the top of each other. The output of lower \lstm forms input to the upper \lstm. For example, if $h_t$ is the output of lower \lstm, then we apply a matrix transform to form the input $x_t$ for the upper \lstm. The Matrix transformation enables us to have two consecutive \lstm layers of different sizes.

\subsubsection{GRU}
\gru is quite a similar network to the \lstm, without any memory unit. \gru network also uses a different gating mechanism with $reset$ ($r$) and $update$ ($z$) gates. The following set of equations defines a \gru model:
\begin{flalign}
& r_t = sigm(W_{xr}x_t + W_{hr}h_{t-1} + b_r) \\
& z_t = sigm(W_{xz}x_t + W_{hz}h_{t-1} + b_z) \notag \\
& \widetilde{h}_t = tanh(W_{xh}x_t + W_{hh}(r_t \odot h_{t-1}) + b_h) \notag \\
& h_t = z_t \odot h_{t-1} + (1 - z_t) \odot \widetilde{h}_t \notag \\
& y_t = softmax(W_{hy}h_t + b_y) \notag
\end{flalign}

\subsection{Implementation}
\label{subsec:impl}
All the models were implemented using \footnotemark THEANO \footnotetext{\url{http://deeplearning.net/software/theano/\#download}} framework~\cite{Bergstra:2010,Bastien:2012}. For all the models, the word embedding dimensionality was 100, no of hidden units were 100 and the context word window size was 5 ($w_{i-2}w_{i-1}w_iw_{i+1}w_{i+2}$). We initialized all the square weight matrices as random orthogonal matrices. All the bias vectors were initialized to zero. Other weight matrices were sampled from a Gaussian distribution with mean 0 and variance $0.0001$.

We trained all the models using Truncated Back-Propagation-Through-Time (T-BPTT)~\cite{Werbos:1990} with the stochastic gradient descent. Standard values of hyper-parameters  were used for RNN model training, as suggested in the literature~\cite{Yao:2014,patel:rnn-qe:2016}. The depth of BPTT was fixed to 7 for all the models. We trained each model for 50 epochs and used Ada-delta~\cite{Zeiler:2002} to adapt the learning rate of each parameter automatically ($\epsilon = 10^{-6}$ and $\rho = 0.95$).

\subsection{Data} \label{sec:data}
We used the data shared by the contest organizers~\cite{jamatia:2016}. The code-mixed data of \bnen, \hien and \teen was shared separately for the Facebook (\fb), Twitter (\twt) and Whatsapp (\wa) posts and conversations with Coarse-Grained (CG) and Fine-Grained (FG) POS annotations. We combined the data from \fb, \twt, and \wa for CG and FG annotation of each language pair. The data was divided into training, testing, and development sets. Testing and development sets were randomly sampled from the complete data. Table~\ref{tab:data} details sizes of the different sets at the sentence and token level. Tag-set counts for CG and FG are also provided.

We preprocess the text for Mentions, Hashtags, Smilies, URLs, Numbers and, Punctuations. In the preprocessing, we mapped all the words of a group to a single new token as they have the same POS tag. For example, all the Mentions like @dhoni, @bcci, and @iitb were mapped to @user; all the Hashtags like \#dhoni, \#bcci, \#iitb were mapped to \#user.

\begin{table*}[hbt]
	\centering
	\begin{tabular}{l|ccc|ccc|cc}
		\multicolumn{1}{c|}{} & \multicolumn{3}{c|}{\#sentences} & \multicolumn{3}{c|}{\#tokens} & \multicolumn{2}{c}{\#tags} \\
		\multicolumn{1}{l|}{code-mix} & \multicolumn{1}{l}{training} & \multicolumn{1}{l}{dev} & \multicolumn{1}{l|}{testing} &  \multicolumn{1}{l}{training} & \multicolumn{1}{l}{dev} & \multicolumn{1}{l|}{testing} &  \multicolumn{1}{l}{CG} & \multicolumn{1}{l}{FG} \\ \hline
		\multicolumn{1}{c|}{\hien} & 2430 & 100 & 100 & 37799 & 1888 & 1457 & 18 & 40 \\
		\multicolumn{1}{c|}{\bnen} & 524 & 50 & 50 & 11977 & 1477 & 1231 & 18 & 38 \\
		\multicolumn{1}{c|}{\teen} & 1779 & 100 & 100 & 26470 & 1436 & 1543 & 18 & 50 \\
	\end{tabular}
	\caption{Data Distribution; CG: Coarse-Grained, FG: Fine-Grained}
	\label{tab:data}
\end{table*}

\subsection{Methodology} \label{sec:methodology}
The \rnnlm models use only the context words' embedding as the input features. We experimented with three \rnn model configurations. In the first setting (\srnn, \lstm, \dlstm, \gru), we learn the word representation from scratch with the other model parameters. In the second configuration (\grup), we trained word representations (pre-training) using $word2vec$~\cite{Mikolov:MT1:2013} tool and fine tuned with the training of other parameters of the network. Pre-training not only guides the learning towards minima with better generalization in non-convex optimization~\cite{bengio:2009,erhan:2010} but also improves the accuracy of the system~\cite{kreutzer:2015,patel:rnn-qe:2016}. In the third setting (\grupl), we also added language of the words as an additional feature with the context words. We learn the vector representation of languages similar to that of words, from scratch.

\section{Results}
We used F1-Score to evaluate the experiments, results are displayed in the Table~\ref{tab:results}. We trained models as described in the section~\ref{sec:methodology}. To compare our results, we also trained the \stan and \hun taggers on the same data, accuracy is given in Table~\ref{tab:results}.

From the table, it is evident that pre-training and language as an additional feature is helpful. Also, the accuracy of our best system (\grupl) is comparable to that of \stan and \hun. \gru models are out-performing other models (\srnn, \lstm, \dlstm) for this task also as reported by~\newcite{Chung:2014} for a suit of NLP tasks.

\begin{table*}[hbt]
	\centering
	\begin{tabular}{l|cc|cc|cc}
		\multicolumn{1}{l|}{} & \multicolumn{2}{c|}{\hien \%F1 score} & \multicolumn{2}{c|}{\bnen \%F1 score} & \multicolumn{2}{c}{\teen \%F1 score} \\
		\multicolumn{1}{l|}{model} & \multicolumn{1}{c}{CG} & \multicolumn{1}{c|}{FG} & \multicolumn{1}{c}{CG} & \multicolumn{1}{c|}{FG} & \multicolumn{1}{c}{CG} & \multicolumn{1}{c}{FG} \\ \hline
		\multicolumn{1}{l|}{\srnn} & 78.16 & 68.73 & 70.16 & 64.49 & 72.27 & \bf 69.04 \\
		\multicolumn{1}{l|}{\lstm} & 62.75 & 53.94 & 41.91 & 35.05 & 57.59 & 51.45 \\
		\multicolumn{1}{l|}{\dlstm} & 70.07 & 59.78 & 54.64 & 46.88 & 65.86 & 59.45 \\
		\multicolumn{1}{l|}{\gru} & \bf 78.29 & \bf 69.32 & \bf 71.90 & \bf 64.96 & \bf 72.40 & 68.72 \\ \hline
		\multicolumn{1}{l|}{\grup} & 80.51 & 71.72 & 74.77 & 68.54 & 74.02 & 70.05 \\
		\multicolumn{1}{l|}{\grupl} & 80.92 & 73.10 & 74.05 & 69.23 & 74.00 & 70.33 \\ \hline
		\multicolumn{1}{l|}{\hun} & 77.50 & 69.04 & 76.55 & 71.02 & 74.30 & 70.73 \\
		\multicolumn{1}{l|}{\stan} & 79.89 & 73.91 & 79.36 & 73.44 & 77.05 & 73.42
	\end{tabular}
	\caption{F1 scores for different experiments}
	\label{tab:results}
\end{table*}

\section{Submission to the Shared Task} \label{sec:submission}
The contest was having two type of submissions, first, $constrained$: restricted to use only the data shared by the organizers with the participants' implemented systems; second, $unconstrained$: participants were allowed to use the publicly available resources (training data, implemented systems etc.).

We submitted for all the language pairs (\hien, \bnen and, \teen) and domains (\fb, \twt and, \wa). For constrained submission, the output of \grupl was used. We trained \stan POS tagger with the same data for $unconstrained$ submission.~\newcite{jamatia:2016} evaluated all the submitted systems against another gold-test set and reported the results.



\section{Analysis}
We did a preliminary analysis of our systems and reported few points in this section. 
\begin{itemize}
	\item The POS categories, contributing more in the error are G\_X, G\_V, G\_N and G\_J for coarse-grained and V\_VM, JJ, N\_NN and N\_NNP for fine-grained systems. Also, we did the confusion matrix analysis and found that these POS tags are mostly confused with each other only. For instance, G\_J POS tag was tagged 28 times wrongly to the other POS tags in which 17 times it was G\_N. 
	
	\item \rnn models require a huge amount of corpus to train the model parameters. From the results, we can observe that for \hien and \teen with only approx 2K training sentences, the results of best \rnn model (\grupl) are comparable to \stan and \hun. For \bnen, the corpus was very less (only approx 0.5K sentences) for \rnn training which resulted into poor performance compared to \stan and \hun. With this and the earlier work on RNN based POS tagging, we can expect that \rnn models could achieve state-of-the-art accuracy with given the sufficient amount of training data. 
	
	\item In general, \lstm and \dlstm models perform better than \srnn. But here, \srnn is outperforming both \lstm and \dlstm. The reason could be less amount of data for training such a complex model. 
	
	\item Few orthographically similar words of English and Hindi, having different POS tags are given with examples in Table~\ref{tab:word-list}. System confuses in POS tagging of such words. With adding language as an additional feature, we were able to tag these type of words correctly.
	
\end{itemize}

\begin{table}[!hbt]
	\centering
	\begin{tabular}{l|c|l|c}
		word & lang &  example & POS \\ \hline
		are & \hi & \textbf{are} shyaam kidhar ho? & PSP \\
		are & \en & they \textbf{are} going. & G\_V \\ \hline
		to & \hi & tumane \textbf{to} dekha hi nhi. & G\_PRT \\
		to & \en & they go \textbf{to} school. & CC \\ \hline
		hi & \hi & mummy to aisi \textbf{hi} hain. & G\_V \\
		hi & \en & \textbf{hi}, how are you. & G\_PRT
	\end{tabular}
	\caption{Similar words in \hien data}
	\label{tab:word-list}
\end{table}

\section{Conclusion and Future Work}
We developed language independent and generic POS tagger for social media text using \rnn networks. We tried \srnn, \lstm, \dlstm and, \gru models. We showed that \gru outperforms other models, and also benefits from pre-training and language as an additional feature. Also, the accuracy of our approach is comparable to that of \stan and \hun.

In the future, we could try \rnn models with more features like POS tags of context words, prefixes and suffixes, length, position, etc. Word characters also have been found to be a very useful feature in \rnn based POS taggers.

\bibliography{eacl2017}
\bibliographystyle{eacl2017}

\end{document}